\documentclass[11pt,a4paper]{article}
\usepackage[hyperref]{acl2021}
\usepackage{times}
\usepackage{latexsym}
\usepackage{amsmath}
\usepackage{mathtools}
\usepackage{amssymb}
\usepackage{array}
\usepackage{amsthm}
\usepackage{multirow}
\usepackage{bbold}
\usepackage{cleveref}
\usepackage{subcaption}
\usepackage{booktabs}
\usepackage{enumitem}
\usepackage{graphicx}

\usepackage{microtype}
\usepackage{soul}

\aclfinalcopy 
\def\aclpaperid{2409} 

\title{Adaptive Nearest Neighbor Machine Translation}

\author{
  Xin Zheng$^1$, Zhirui Zhang$^2$, Junliang Guo$^3$, Shujian Huang$^1$\thanks{~ ~Corresponding author.}, Boxing Chen$^2$, \\ 
  {\bf Weihua Luo$^2$ and Jiajun Chen$^1$} \\
  $^1$National Key Laboratory for Novel Software Technology, Nanjing University, China \\
  $^2$Machine Intelligence Technology Lab, Alibaba DAMO Academy \\
  $^3$University of Science and Technology of China \\
  $^1$\texttt{zhengxin@smail.nju.edu.cn},\; \texttt{\{huangsj,chenjj\}@nju.edu.cn}\\ 
  $^2$\texttt{\{zhirui.zzr, boxing.cbx, weihua.luowh\}@alibaba-inc.com}\\
  $^3$\texttt{guojunll@mail.ustc.edu.cn}
}

\date{}
\begin{document}
\maketitle
\begin{abstract}
$k$NN-MT, recently proposed by \citet{DBLP:journals/corr/abs-2010-00710}, successfully combines pre-trained neural machine translation (NMT) model with token-level $k$-nearest-neighbor ($k$NN) retrieval to improve the translation accuracy.
However, the traditional $k$NN algorithm used in $k$NN-MT 
simply retrieves a same number of nearest neighbors for each target token,
which may cause prediction errors when the retrieved neighbors include noises.
In this paper, we propose Adaptive $k$NN-MT to dynamically determine the number of $k$ for each target token. We achieve this by introducing a light-weight \textit{Meta-$k$ Network},
which can be efficiently trained with only a few training samples.
On four benchmark machine translation datasets, we demonstrate that the proposed method is able to effectively filter out the noises in retrieval results and significantly outperforms the vanilla $k$NN-MT model.
Even more noteworthy is that the Meta-$k$ Network learned on one domain could be directly applied to other domains and obtain consistent improvements, illustrating the generality of our method. 
Our implementation is open-sourced at \url{https://github.com/zhengxxn/adaptive-knn-mt}.
\end{abstract}

\section{Introduction}
\begin{figure*}[t]
    \centering
    \includegraphics[width=0.94\textwidth]{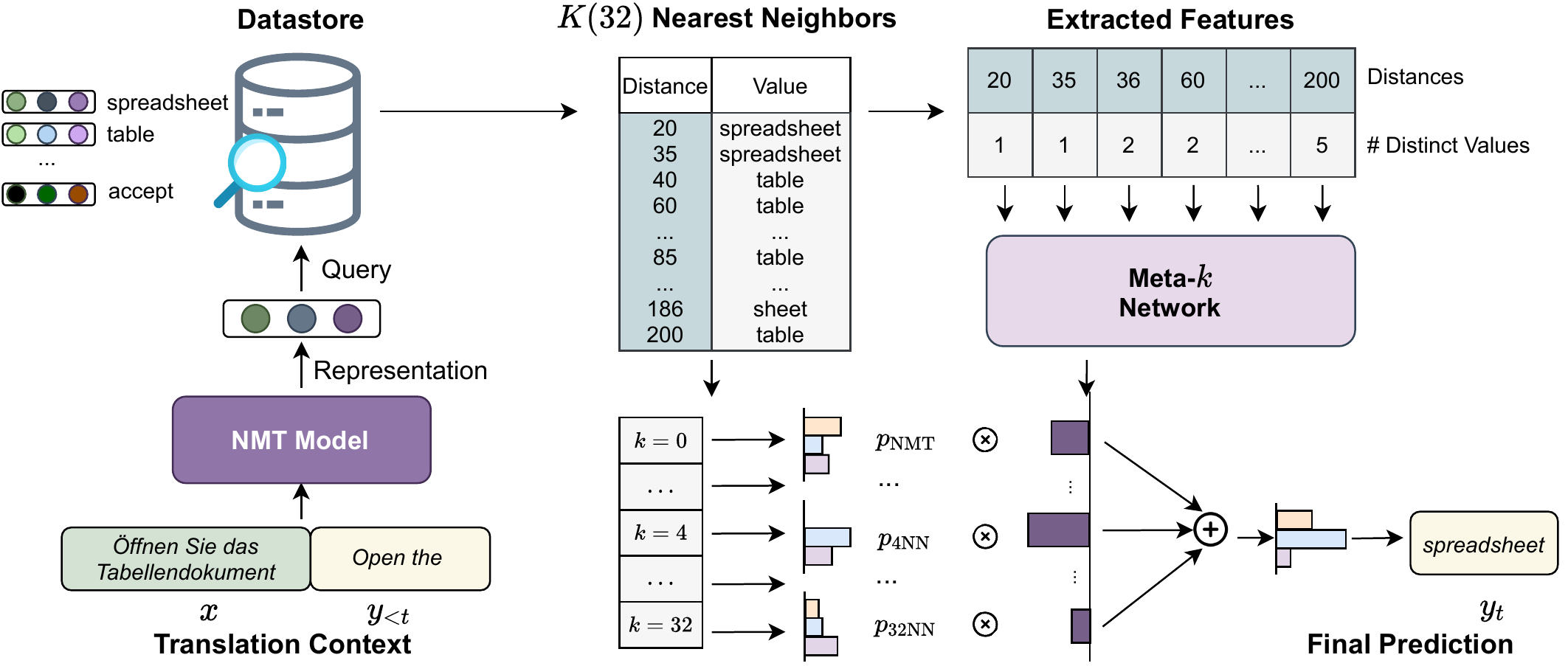}
    \caption{An overview of the proposed Adaptive $k$NN-MT, which could dynamically evaluate and aggregate a set of $k$NN predictions based on the distances as well as count of distinct values of retrieved neighbors.}
    \label{fig:model}
    \vspace{-3pt}
\end{figure*}

Retrieval-based methods \citep{DBLP:conf/aaai/GuWCL18, zhang-etal-2018-guiding, bapna-firat-2019-non, DBLP:journals/corr/abs-2010-00710} are increasingly receiving attentions from the machine translation (MT) community recently.
These approaches complement advanced neural machine translation (NMT) models \citep{DBLP:conf/nips/SutskeverVL14,DBLP:journals/corr/BahdanauCB14,DBLP:conf/nips/VaswaniSPUJGKP17,Hassan2018AchievingHP} to alleviate the performance degradation when translating out-of-domain sentences~\cite{dou-etal-2019-unsupervised,wei-emnlp-2020-iterative}, rare words~\cite{DBLP:conf/aclnmt/KoehnK17}, etc.
The ability of accessing any provided datastore during translation makes them scalable, adaptable and interpretable.

$k$NN-MT, recently proposed in~\citep{DBLP:journals/corr/abs-2010-00710}, equips a pre-trained NMT model with a $k$NN classifier over a datastore of cached context representations and corresponding target tokens, providing a simple yet effective strategy to utilize cached contextual information in inference.
However, the hyper-parameter $k$ is fixed for all cases, which raises some potential problems.
Intuitively, the retrieved neighbors may include noises when the target token is relatively hard to determine (e.g., relevant context is not enough in the datastore).
And empirically, we find that the translation quality is very sensitive to the choice of $k$, results in the poor robustness and generalization performance.


To tackle this problem, we propose Adaptive $k$NN-MT that determines the choice of $k$ regarding each target token adaptively. Specifically, instead of utilizing a fixed $k$, we consider a set of possible $k$ 
that are smaller than an upper bound $K$. 
Then, given the retrieval results of the current target token, we propose a light-weight \textit{Meta-$k$ Network} to estimate the importance of all possible $k$-Nearest Neighbor results, based on which they are aggregated to obtain the final decision of the model.
In this way, our method dynamically evaluate and utilize the neighbor information conditioned on different target tokens, therefore improve the translation performance of the model.

We conduct experiments on multi-domain machine translation datasets.
Across four domains, our approach can achieve 1.44$\sim$2.97 BLEU score improvements over the vanilla $k$NN-MT on average when $K\geq4$.
The introduced light-weight Meta-$k$ Network only requires thousands of parameters and can be easily trained with a few training samples.
In addition, we find that the Meta-$k$ Network trained on one domain can be directly applied to other domains and obtain strong performance, showing the generality and robustness of the proposed method.



\section{Background: $k$NN-MT}

In this section, we will briefly introduce the background of $k$NN-MT, which includes two steps: creating a datastore and making predictions depends on it.

\paragraph{Datastore Creation.} 
The datastore consists of a set of key-value pairs. 
Formally, given a bilingual sentence pair in the training set $(x, y) \in (\mathcal{X}, \mathcal{Y})$,
a pre-trained autoregressive NMT decoder translates the $t$-th target token $y_t$ based on the translation context $(x, y_{<t})$. 
Denote the hidden representations of translation contexts as $f(x, y_{<t})$, then the datastore is constructed by taking $f(x, y_{<t})$ as keys and $y_t$ as values, 
\begin{equation*}
(\mathcal{K}, \mathcal{V}) = \bigcup_{(x, y) \in (\mathcal{X}, \mathcal{Y})} \{(f(x, y_{<t}), y_t), \forall y_t \in y \}. 
\label{equ:datastore}
\end{equation*}
Therefore, the datastore can be created through a single forward pass over the training set $(\mathcal{X}, \mathcal{Y})$.

\paragraph{Prediction.}
While inference, at each decoding step $t$, the $k$NN-MT model aims to predict $\hat{y}_t$ given the already generated tokens $\hat{y}_{<t}$ as well as the context representation $f(x, \hat{y}_{<t})$, which is utilized to query the datastore for $k$ nearest neighbors w.r.t the $l_2$ distance. Denote the retrieved neighbors as $N^t = \{(h_i, v_i), i \in \{1, 2, ..., k\} \}$, their distribution over the vocabulary is computed as:
\begin{align}
p_{\textrm{kNN}}(y_t & |x, \hat{y}_{<t}) \propto \label{equ:knn_prob} \\
&\sum_{(h_i, v_i)} \mathbb{1}_{y_t = v_i} \exp (\frac{-d(h_i, f(x, \hat{y}_{<t}))}{T}), \nonumber 
\end{align}
where $T$ is the temperature and $d(\cdot, \cdot)$ indicates the $l_2$ distance.
The final probability when predicting $y_t$ is calculated as the interpolation of two distributions with a hyper-parameter $\lambda$:
\begin{equation}
\label{equ:prob_ip}
\begin{split}
p(y_t|x, \hat{y}_{<t}) &= \lambda \ p_{\textrm{kNN}}(y_t|x, \hat{y}_{<t}) \\
& + (1-\lambda) \ p_{\textrm{NMT}} (y_t|x, \hat{y}_{<t}), 
\end{split}
\end{equation}
where $p_{\textrm{NMT}}$ indicates the vanilla NMT prediction.


\section{Adaptive $k$NN-MT}
\label{sec:method}

The vanilla $k$NN-MT method utilizes a fixed number of translation contexts for every target token, which fails to exclude noises contained in retrieved neighbors when there are not enough relevant items in the datastore. 
We show an example with $k=32$ in \Cref{fig:model}. The correct prediction \textit{spreadsheet} has been retrieved as top candidates. However, the model will finally predict \textit{table} instead because it appears more frequently in the datastore than the correct prediction.
A naive way to filter the noises is to use a small $k$, but this will also cause over-fitting problems for other cases. 
In fact, the optimal choice of $k$ varies when utilizing different datastores in vanilla $k$NN-MT, leading to poor robustness and generalizability of the method, which is empirically discussed in \Cref{sec:different k}.


To tackle this problem, we propose a dynamic method that allows each untranslated token to utilize different numbers of neighbors. Specifically, we consider a set of possible $k$s that are smaller than an upper bound $K$, and introduce a light-weight \textit{Meta-$k$ Network} to estimate the importance of utilizing different $k$s. 
Practically, we consider the powers of $2$ as the choices of $k$ for simplicity, as well as $k = 0$ which indicates ignoring $k$NN and only utilizing the NMT model, i.e., $k \in \mathcal{S}$ where $\mathcal{S}=\{0 \} \cup \{ k_i \in \mathbb{N} \mid \log_2 k_i \in \mathbb{N}, k_i \leq K \}$. 
Then the Meta-$k$ Network evaluates the probability of different $k$NN results by taking retrieved neighbors as inputs. 

Concretely, at the $t$-th decoding step, we first retrieve $K$ neighbors $N^{t}$ from the datastore, and for each neighbor $(h_i, v_i)$, we calculate its distance from the current context representation $d_i=d(h_i, f(x, \hat{y}_{<t}))$, as well as the count of distinct values in top $i$ neighbors $c_i$.
Denote $d=(d_1,...,d_K)$ as distances and $c=(c_1,...,c_K)$ as counts of values for all retrieved neighbors, we then concatenate them as the input features to the Meta-$k$ Network. The reasons of doing so are two-fold.
Intuitively, the distance of each neighbor is the most direct evidence when evaluating their importance. 
In addition, the value distribution of retrieved results is also crucial for making the decision, i.e., if the values of each retrieved results are distinct, then the $k$NN predictions are less credible and we should depend more on NMT predictions.

We construct the Meta-$k$ Network $f_{\textrm{Meta}}(\cdot)$ as two feed-forward Networks with non-linearity between them.
Given $[d;c]$ as input, the probability of applying each $k$NN results is computed as:
\begin{equation}
p_{\textrm{Meta}}(k) = \textrm{softmax} (f_{\textrm{Meta}}([d;c])).
\label{equ:prob_estimate}
\end{equation}

\paragraph{Prediction.}
Instead of introducing the hyper-parameter $\lambda$ as 
\Cref{equ:prob_ip},
we aggregate the NMT model and different $k$NN predictions with the output of the Meta-$k$ Network to obtain the final prediction:
\begin{equation}
\label{equ:adaptive_prob}
p(y_t| x,\hat{y}_{<t}) =  \sum_{k_i \in \mathcal{S}} p_{\textrm{Meta}}(k_i) \cdot p_{k_i\textrm{NN}}(y_t|x, \hat{y}_{<t}),
\end{equation}
where $p_{k_i\textrm{NN}}$ indicates the $k_i$ Nearest Neighbor prediction results calculated as \Cref{equ:knn_prob}.


\paragraph{Training.}

We fix the pre-trained NMT model and only optimize the Meta-$k$ Network by minimizing the cross entropy loss following Equation~(\ref{equ:adaptive_prob}), which could be very efficient by only utilizing hundreds of training samples.

\section{Experiments}

\subsection{Experimental Setup}

We evaluate the proposed model in domain adaptation machine translation tasks,
in which a pre-trained general-domain NMT model is used to translate domain-specific sentences with $k$NN searching over an in-domain datastore.
This is the most appealing application of $k$NN-MT as it could achieve comparable results with an in-domain NMT model but without training on any in-domain data.
We denote the proposed model as Adaptive $k$NN-MT (A) and compare it with two baselines. One of that is vanilla $k$NN-MT (V) and the other is uniform $k$NN-MT (U) where we set equal confidence for each $k$NN prediction.


\paragraph{Datasets and Evaluation Metric.} 
We use the same multi-domain dataset as the baseline~\citep{DBLP:journals/corr/abs-2010-00710}, 
and consider domains including \textbf{IT}, \textbf{Medical}, \textbf{Koran}, and \textbf{Law} in our experiments.
The sentence statistics of datasets are illustrated in \Cref{table:dataset}.
The Moses toolkit\footnote{\url{https://github.com/moses-smt/mosesdecoder}} is used to tokenize the sentences and split the words into subword units \citep{sennrich-etal-2016-neural} with the bpe-codes provided by \citet{ng-etal-2019-facebook}.
We use SacreBLEU\footnote{\url{https://github.com/mjpost/sacrebleu}} to measure all results with case-sensitive detokenized BLEU~\citep{papineni-etal-2002-bleu}.

\begin{table}[htb] \small
\centering
\resizebox{0.48\textwidth}{!}{%
\begin{tabular}{l|cccc}
\toprule
\multicolumn{1}{l|}{Dataset} & IT    & Medical & Koran & Laws    \\ \midrule
Train                     & $222,927$ & $248,009$   & $17,982$ & $467,309$ \\ 
Dev            & $2000$& $2000$   & $2000$ & $2000$ \\ 
Test                     & $2000$ & $2000$   & $2000$ & $2000$ \\ \bottomrule
\end{tabular}%
}
\vspace{-5pt}
\caption{Statistics of dataset in different domains.}
\label{table:dataset}
\vspace{-5pt}
\end{table}



\begin{table*}[tb] \small
\centering
\setlength\tabcolsep{2pt}
\begin{tabular}{@{}c|c|ccccccccccccccc@{}}
\toprule
\multicolumn{1}{l|}{} & Domain & \multicolumn{3}{c|}{IT {\tiny (Base NMT: $38.35$)}} & \multicolumn{3}{c|}{Med {\tiny (Base NMT: $39.99$)}} & \multicolumn{3}{c|}{Koran {\tiny (Base NMT: $16.26$)}} & \multicolumn{3}{c|}{Law {\tiny (Base NMT: $45.48$)}} & \multicolumn{3}{c}{Avg {\tiny (Base NMT: $35.02$)}}\\ \midrule
\multicolumn{1}{l|}{} & Model & V & U & \multicolumn{1}{c|}{A} & V & U & \multicolumn{1}{c|}{A} & V & U & \multicolumn{1}{c|}{A} & V & U & \multicolumn{1}{c|}{A} & V & U & A \\ \midrule
\multirow{6}{*}{K}    & $1$      & $42.19$  & $41.21$  & \multicolumn{1}{c|}{$42.52$} & $51.41$ & $50.32$ & \multicolumn{1}{c|}{$51.82$} & $18.12$ & $17.15$ & \multicolumn{1}{c|}{$18.10$} & $58.76$ & $58.05$ & \multicolumn{1}{c|}{$58.81$} & $42.62$ & $41.68$ & $42.81$ \\
                      & $2$      & $44.20$  & $41.43$  & \multicolumn{1}{c|}{$46.18$} & $53.65$ & $52.44$ & \multicolumn{1}{c|}{$55.20$} & $19.37$ & $17.36$ & \multicolumn{1}{c|}{$19.12$} & $60.80$ & $59.81$ & \multicolumn{1}{c|}{$61.76$} & $44.50$ & $42.76$ & $45.56$ \\
                      & $4$      & $44.89$  & $42.31$  & \multicolumn{1}{c|}{$47.23$} & \underline{$54.16$} & $53.01$ & \multicolumn{1}{c|}{$55.84$} & $19.50$ & $17.88$ & \multicolumn{1}{c|}{$19.69$} & \underline{$61.31$} & $60.75$ & \multicolumn{1}{c|}{$62.89$} & $44.97$ & $43.49$ & $46.41$ \\
                      & $8$      & \underline{$45.96$}  & $42.46$  & \multicolumn{1}{c|}{$\mathbf{48.04}$} & $54.06$ & $53.46$ & \multicolumn{1}{c|}{$56.31$} & $20.12$ & $18.59$ & \multicolumn{1}{c|}{$20.57$} & $61.12$ & $61.37$ & \multicolumn{1}{c|}{$\mathbf{63.21}$} & \underline{$45.32$} & $43.97$ & $47.03$ \\
                      & $16$     & $45.36$  & $43.05$  & \multicolumn{1}{c|}{$47.71$} & $53.54$ & \underline{$54.08$} & \multicolumn{1}{c|}{$\mathbf{56.41}$} & \underline{$20.30$} & $19.45$ & \multicolumn{1}{c|}{$\mathbf{21.09}$} & $60.21$ & $61.52$ & \multicolumn{1}{c|}{$63.07$} & $44.85$ & $44.53$ & $\mathbf{47.07}$ \\
                      & $32$     & $44.81$  & \underline{$43.78$}  & \multicolumn{1}{c|}{$47.68$} & $52.52$ & $53.95$ & \multicolumn{1}{c|}{$56.21$} & $19.66$ & \underline{$19.99$} & \multicolumn{1}{c|}{$20.96$} & $59.04$ & \underline{$61.53$} & \multicolumn{1}{c|}{$63.03$} & $44.00$ & \underline{$44.81$} & $46.97$ \\ \midrule
\multicolumn{1}{l|}{} & $\sigma^2${\tiny($K \geq 4 $)} & $0.21$  & $0.33$& \multicolumn{1}{c|}{$\mathbf{0.08}$} & $0.42$ & $0.18$ & \multicolumn{1}{c|}{$\mathbf{0.05}$} & $\mathbf{0.10}$ & $0.65$ & \multicolumn{1}{c|}{$0.30$} & $0.81$ & $0.10$ &  \multicolumn{1}{c|}{$\mathbf{0.01}$} & $0.24$ & $0.26$ & $\mathbf{0.07}$ \\ \bottomrule
\end{tabular}%
\vspace{-5pt}
\caption{The BLEU scores of the vanilla $k$NN-MT (V) and uniform $k$NN-MT (U) baselines and the proposed Adaptive $k$NN-MT model (A). Underline results indicate the best results of baselines, and our best results are marked bold. $\sigma^2$ indicates the variance of results among different $K$s.
}
\label{table:main}
\vspace{-5pt}
\end{table*}

\paragraph{Implementation Details.}

We adopt the fairseq toolkit\footnote{\url{https://github.com/pytorch/fairseq}}\cite{ott-etal-2019-fairseq} and faiss\footnote{\url{https://github.com/facebookresearch/faiss}}\citep{DBLP:journals/corr/JohnsonDJ17} to replicate $k$NN-MT and implement our model.
We apply the WMT'19 German-English news translation task winner model \citep{ng-etal-2019-facebook} as the pre-trained NMT model which is also used by \citet{DBLP:journals/corr/abs-2010-00710}.
For $k$NN-MT, we carefully tune the hyper-parameter $\lambda$ in \Cref{equ:prob_ip} and report the best scores for each domain.
More details are included in the supplementary materials.
For our method, the hidden size of the two-layer FFN in Meta-$k$ Network is set to $32$.
We directly use the dev set (about $2$k sents) to train the Meta-$k$ Network
for about 5k steps.
We use Adam \citep{DBLP:journals/corr/KingmaB14} to optimize our model, the learning rate is set to 3e-4 and batch size is set to $32$ sentences.

\subsection{Main Results}

\begin{table}[tb]
\centering
\resizebox{0.48\textwidth}{!}{%
\begin{tabular}{l|cccc|c}
\toprule
\multicolumn{1}{l|}{Adaptive $k$NN-MT} & IT    & Medical & Koran & Law  & Avg   \\ \midrule
In-domain                   & $47.68$ & $56.21$   & $20.96$ & $63.03$ & $46.97$ \\ 
IT domain            & $47.68$ & $56.20$   & $20.52$ & $62.33$ & $46.68$ \\ \bottomrule
\end{tabular}%
}
\vspace{-5pt}
\caption{Generality Evaluation. We train the model on the IT domain and directly apply to other test sets.}
\label{table:generalization}
\vspace{-5pt}
\end{table}

\begin{table}[tb]
\centering
\small
\begin{tabular}{l|cc}
\toprule
\multicolumn{1}{l|}{Model}  & IT $\Rightarrow$ Medical & Medical $\Rightarrow$ IT \\ \midrule
Base NMT                     & $39.99$   & $38.35$ \\ \midrule
$k$NN-MT                     & $25.82$   & $15.79$        \\ 
Adaptive $k$NN-MT            & $37.78$   & $30.09$          \\ \bottomrule
\end{tabular}%
\vspace{-5pt}
\caption{Robustness Evaluation, where the test sets are from Medical/IT domains and the datastore are from IT/Medical domains respectivally.} 
\label{table:out-domain}
\vspace{-5pt}
\end{table}

The experimental results are listed in~\Cref{table:main}.
We can observe that the proposed Adaptive $k$NN-MT significantly outperforms the vanilla $k$NN-MT on all domains, illustrating the benefits of dynamically determining and utilizing the neighbor information for each target token.
In addition, the performance of the vanilla model is sensitive to the choice of $K$, while our proposed model is more robust with smaller variance.
More specifically, our model achieves better results when choosing larger number of neighbors, while the vanilla model suffers from the performance degradation when $K=32$, indicating that the proposed Meta-$k$ Network is able to effectively evaluate and filter the noise in retrieved neighbors, while a fixed $K$ cannot.
We also compare our proposed method with another naive baseline, uniform $k$NN-MT, where we set equal confidence for each $k$NN prediction and make it close to the vanilla $k$NN-MT with small $k$. 
It further demonstrates that our method could really learn something useful but not bias smaller $k$.

\paragraph{Generality.}
To demonstrate the generality of our method, we directly utilize the  Meta-$k$ Network trained on the IT domain to evaluate other domains. 
For example, we use the Meta-$k$ Network trained on IT domain and medical datastore to evaluate the performance on medical test set.
For comparison, we collect the in-domain results from \Cref{table:main}. We set $K=32$ for both settings.
As shown in \Cref{table:generalization}, the Meta-$k$ Network trained on the IT domain achieves comparable performance on all other domains which re-train the Meta-$k$ Network with in-domain dataset.
These results also indicate that the mapping from our designed feature to the confidence of retrieved neighbors is common across different domains.

\paragraph{Robustness.}
We also evaluate the robustness of our method in the domain-mismatch setting, where we consider a scenario that the user inputs an out-of-domain sentence (e.g. IT domain) to a domain-specific translation system (e.g. medical domain) to evaluate the robustness of different methods.
Specifically, in IT $\Rightarrow$ Medical setting, we firstly use medical dev set and datastore to tune hyper-parameter for vanilla $k$NN-MT or train the Meta-$k$ Network for Adaptive $k$NN-MT, and then use IT test set to test the model with medical datastore.
We set $K=32$ in this experiment.
As shown in \Cref{table:out-domain}, the retrieved results are highly noisy so that the vanilla $k$NN-MT encounters drastic performance degradation.
In contrast, our method could effectively filter out noises and therefore prevent performance degradation as much as possible.

\begin{table*}[tb] \small
\centering
\begin{tabular}{@{}r|l@{}}
\toprule
Source           & Wenn eine gleichzeitige Behandlung mit Vitamin K Antagonisten erforderlich ist, müssen die  \\ 
& Angaben in Abschnitt 4.5 beachtet werden. \\ \midrule
Reference       & therapy with vitamin K antagonist should be administered in accordance with the information \\
&  of Section 4.5. \\ \midrule
 Base NMT       &  If a simultaneous treatment with vitamin K antagonists is required, the information in section \\
 & 4.5 must be observed. \\ \midrule
$k$NN-MT          & If concomitant treatment with vitamin K antagonists is required, please refer to section 4.5. \\ \midrule
Adaptive $k$NN-MT & When required, concomitant \textbf{\textit{therapy with vitamin K antagonist should be administered in}}  \\
 & \textbf{\textit{accordance with the information of Section 4.5.}} \\ \bottomrule
\end{tabular}%
\vspace{-5pt}
\caption{Translation examples of different systems in Medical domain.}
\label{table:case_study}
\vspace{-5pt}
\end{table*}

\begin{table}[tb]
\centering
\small
\begin{tabular}{@{}r|c@{}}
\toprule
\multicolumn{1}{l|}{Adaptive $k$NN-MT ($K$ = $8$)} & $48.04$ \\ \midrule
- value count feature                         & $46.76$ \\ \midrule
- distance feature                            & $45.60$ \\ \bottomrule
\end{tabular}%
\vspace{-5pt}
\caption{Effect of different features in Meta-$k$ Network.} 
\label{table:feature_ab}
\vspace{-5pt}
\end{table}

\begin{figure}[tb]
    \centering
    \includegraphics[width=0.48\textwidth]{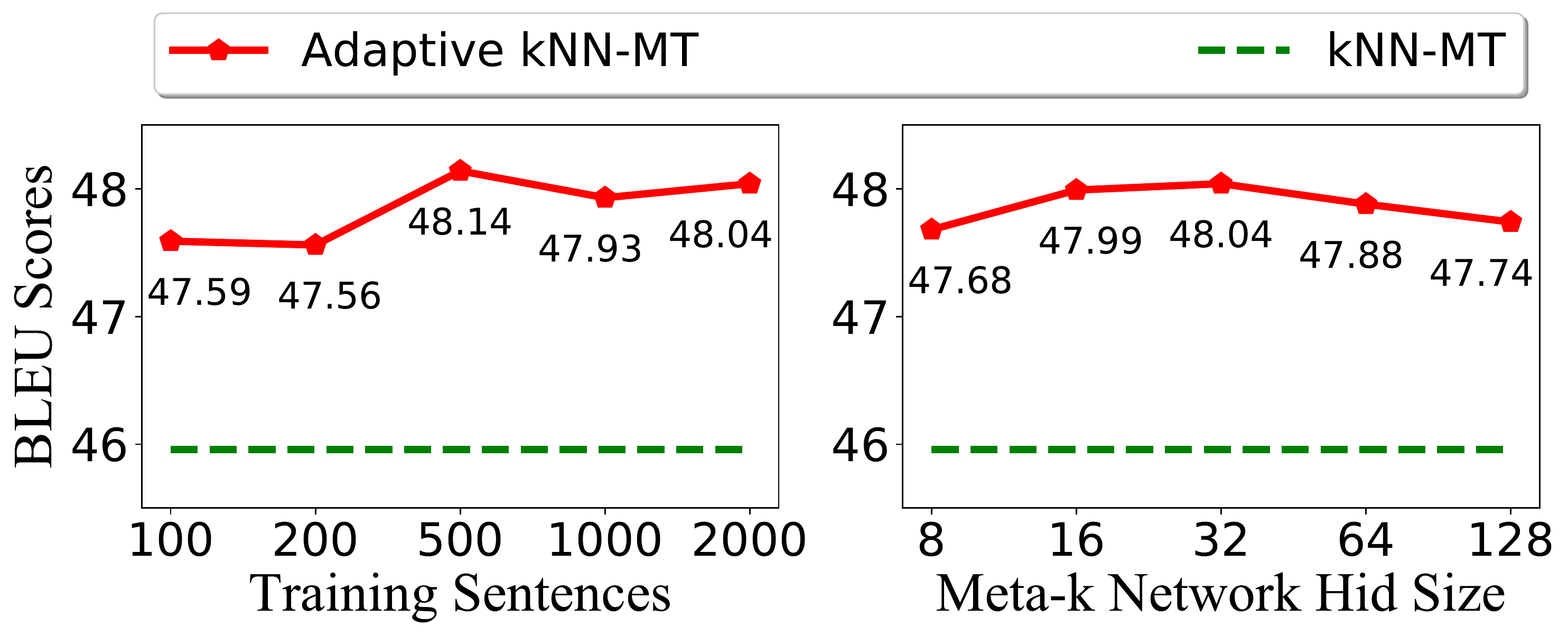}
    \vspace{-10pt}
    \caption{Effect of different number of training sentences and the hidden size of Meta-$k$ Network.}
    \label{fig:analysis}
    \vspace{-10pt}
\end{figure}

\label{sec:different k}

\paragraph{Case Study.} \Cref{table:case_study} shows a translation example selected from the test set in \textbf{Medical} domain with $K=32$. 
We can observe that the Meta-$k$ Network could determine the choice of $k$ for each target token respectively, based on which Adaptive $k$NN-MT  leverages in-domain datastore better to achieve proper word selection and language style.

\paragraph{Analysis.}
Finally, we study the effect of two designed features, number of training sentences and the hidden size of the proposed Meta-$k$ Network.
We conduct these ablation study on IT domain with $K=8$.
All experimental results are summarized in \Cref{table:feature_ab} and  \Cref{fig:analysis}.
It's obvious that both of the two features contribute significantly to the excellent performance of our model, in which the distance feature is more important. 
And surprisingly, our model could outperforms the vanilla $k$NN-MT with only $100$ training sentences, or with a hidden size of $8$ that only contains around $0.6$k parameters, showing the efficiency of our model.



\section{Conclusion and Future Works}
In this paper, we propose Adaptive $k$NN-MT model to dynamically determine the utilization of retrieved neighbors for each target token, by introducing a light-weight \textit{Meta-$k$ Network}.
In the experiments, on the domain adaptation machine translation tasks, we demonstrate that our model is able to effectively filter the noises in retrieved neighbors and significantly outperform the vanilla $k$NN-MT baseline.
In addition, the superiority of our method on generality and robustness is also verified.
In the future, we plan to extend our method to other tasks like Language Modeling, Question Answering, etc, which can also benefit from utilizing $k$NN searching \citep{DBLP:conf/iclr/KhandelwalLJZL20,kassner-schutze-2020-bert}.

\section{Acknowledgments}
We would like to thank the anonymous reviewers for the helpful comments.
This work was supported by the National Key R\&D Program of China (No. 2019QY1806),
National Science Foundation of China (No. 61772261, U1836221) and Alibaba Group through Alibaba Innovative Research Program.
We appreciate Weizhi Wang, Hao-Ran Wei and Jun Xie for the fruitful discussions. 
The work was done when the first author was an intern at Alibaba Group.

\bibliographystyle{acl_natbib}
\bibliography{,acl2021}

\appendix

\section{Appendix}

\begin{table*}[htb] 
\small
\centering
\begin{tabular}{l|cccc}
\toprule
\multicolumn{1}{l|}{Datastore} & IT    & Medical & Koran & Laws    \\ \midrule
Size                     & $3,613,350$ & $6,903,320$   & $524,400$ & $19,070,000$ \\ 
Hard Disk Space (Datastore)            & $6.9$ Gb& $15$ Gb   & $1.1$ Gb & $37$ Gb \\ 
Hard Disk Space (faiss index)                     & $266$ Mb & $492$ Mb   & $54$ Mb & $1.3$ Gb \\ \bottomrule
\end{tabular}%
\vspace{-5pt}
\caption{Statistics of datastore in different domains.}
\label{table:disk space}
\vspace{-5pt}
\end{table*}

\begin{table*}[htb] 
\small
\centering
\begin{tabular}{l|c|cccc}
\toprule
ms / sent & $K$     & Batch=1 & Batch=16 & Batch=32 & Batch=64   \\ 
\midrule
NMT  & $0$     & $165$ & $16.3$ & $10.5$   & $7.9$ \\ \midrule
\multirow{3}{*}{$k$NN-MT}  
 & $8$   & $291.0 (\times 1.76)$ & $51.0 (\times 3.1)$ & $43.6 (\times 4.2)$   & $38.0 (\times 4.8)$  \\
 & $16$  & $311.4 ( \times 1.89) $ & $81.1 (\times 5.0)$ & $70.4 (\times 6.7)$   & $64.4 (\times 8.2)$ \\
 & $32$  & $385.5 (\times 2.34) $ & $136.5 (\times 8.4)$ & $123.8 (\times 11.8)$   & $114.9 (\times 14.5)$  \\
\midrule
\multirow{3}{*}{Adaptive $k$NN-MT}
 & $8$   & $299.1 (\times 1.81)$ & $51.1(\times 3.1)$ & $42.8 (\times 4.1)$   & $38.1 (\times 4.8)$ \\
 & $16$  & $315.0 ( \times 1.91) $ & $80.2(\times 4.9)$ & $70.2 (\times 6.7)$   & $63.7 (\times 8.1)$  \\
 & $32$  & $394.5 (\times 2.40) $ & $147.5 (\times 9.0)$ & $128.0 (\times 12.2) $   & $116.8 (\times 14.8)$ \\
\bottomrule
\end{tabular}
\vspace{-5pt}
\caption{Decoding time of different models.
All results are tested on 20 cores Intel(R) Xeon(R) Gold 6248 CPU @ 2.50GHz with a V100-32GB GPU.}
\label{table:decoding time}
\vspace{-5pt}
\end{table*}

\subsection{Datastore Creation}

We first use numpy array to save the key-value pairs over training sets as datastore. 
Then, faiss is used to build index for each datastore to carry out fast nearest neighbor search.
We utilize faiss to learn $4k$ cluster centroids for each domain, and search $32$ clusters for each target token in decoding.
The size of datastore (count of target tokens), and hard disk space of datastore as well as faiss index are shown in \Cref{table:disk space}.

\subsection{Hyper-Parameter Tuning for $k$NN-MT}

The performance of vanilla $k$NN-MT is highly related to the choice of hyper-parameter, i.e. $k$, $T$ and $\lambda$.
We fix $T$ as $10$ for IT, Medical, Law, and $100$ for Koran in all experiments.
Then, we tuned $k$ and $\lambda$ for each domain when using $k$NN-MT and the optimal choice for each domain are shown in \Cref{table:hp-each-domain}.
The performance of $k$NN-MT is unstable with different hyper-parameters while our Adaptive $k$NN-MT avoids this problem.

\begin{table}[ht] 
\centering
\small
\begin{tabular}{c|cccc}
\toprule
\multicolumn{1}{c|}{Dataset} & IT    & Medical & Koran & Laws    \\ \midrule
$k$                    & $8$ & $4$   & $16$ & $4$ \\ 
$T$                    & $10$ & $10$   & $10$ & $100$ \\ 
$\lambda$                    & $0.7$ & $0.8$   & $0.8$ & $0.8$ \\ \bottomrule
\end{tabular}%
\vspace{-5pt}
\caption{Optimal choice of hyper-parameters for each domain in vanilla $k$NN-MT.}
\label{table:hp-each-domain}
\vspace{-5pt}
\end{table}

\subsection{Decoding Time}

We compare the decoding time on IT test set of NMT, $k$NN-MT (our replicated) and Adaptive $k$NN-MT condition on different batch size.
In decoding, the beam size is set to 4 with length penalty 0.6.
The results are summarized in \Cref{table:decoding time}.



\end{document}